\documentclass[10pt,twocolumn,letterpaper]{article}

\usepackage{iccv}
\usepackage{times}
\usepackage{epsfig}
\usepackage{graphicx}
\usepackage{amsmath}
\usepackage{amssymb}

\usepackage{array}
\usepackage{kotex}
\usepackage{amsmath}
\usepackage{mathtools}
\usepackage{multirow}
\usepackage{subcaption}
\usepackage[normalem]{ulem}
\usepackage{adjustbox}
\usepackage{graphicx}
\usepackage{comment}
\usepackage{amsmath,amssymb}
\usepackage{color}
\usepackage{arydshln}
\usepackage{ulem}
\usepackage[accsupp]{axessibility}
 
\newcolumntype{C}[1]{>{\centering\arraybackslash}p{#1}}
\newcolumntype{M}[1]{>{\centering\arraybackslash}m{#1}}
\newcolumntype{L}[1]{>{\arraybackslash}p{#1}}
\newcommand{\argmax}{\mathop{\mathrm{argmax}}}

\usepackage[pagebackref=true,breaklinks=true,letterpaper=true,colorlinks,bookmarks=false]{hyperref}

\iccvfinalcopy 

\def\iccvPaperID{8969} 

\ificcvfinal\fi

\definecolor{dark_cyan}{RGB}{0,139,139}
\definecolor{green}{RGB}{0,84,0}

\newcommand{\jy}[1]{\textcolor{black}{#1}}

\begin{document}

\title{Training Multi-Object Detector by \\Estimating Bounding Box Distribution for Input Image}

\author{
    Jaeyoung Yoo$^{1}$ \qquad 
    Hojun Lee$^{1,*}$ \qquad 
    Inseop Chung$^{1,*}$ \qquad
    Geonseok Seo$^{2}$ \qquad 
    Nojun Kwak$^{1}$ \\
    
    $^{1}$Seoul National University \qquad
    $^{2}$Samsung Advanced Institute of Technology \\
    
    {
         \tt\small 
         \{yoojy31, hojun815, jis3613, nojunk\}@snu.ac.kr \qquad
         gunsuk.seo@samsung.com
    }
}

\maketitle
\ificcvfinal\thispagestyle{empty}\fi

\begin{abstract}
In multi-object detection using neural networks, the fundamental problem is, ``How should the network learn a variable number of bounding boxes in different input images?''. Previous methods train a multi-object detection network through a procedure that directly assigns the ground truth bounding boxes to the specific locations of the network's output. However, this procedure makes the training of a multi-object detection network too heuristic and complicated. In this paper, we reformulate the multi-object detection task as a problem of density estimation of bounding boxes. Instead of assigning each ground truth to specific locations of network's output, we train a network by estimating the probability density of bounding boxes in an input image using a mixture model. For this purpose, we propose a novel network for object detection called Mixture Density Object Detector (MDOD), and the corresponding objective function for the density-estimation-based training. We applied MDOD to MS COCO dataset. Our proposed method not only deals with multi-object detection problems in a new approach, but also improves detection performances through MDOD.
The code is available: \footnotesize{\url{https://github.com/yoojy31/MDOD}}.
\end{abstract}

\section{Introduction}
{\let\thefootnote\relax\footnotetext{{
\jy{
    $^{*}$H. Lee and I. Chung equally contributed to this work. \\ 
    This work was supported by NRF grant (2021R1A2C3006659) and IITP grant (No. 2021-0-01343), both funded by Korean Government. The authors are also funded by Samsung Electronics.}}
}}

Multi-object detection is the task of finding multiple objects through bounding boxes with class information. Since the breakthrough of the deep neural networks, multi-object detection has been extensively developed in terms of computational efficiency and performance, and is now at a level that can be used in real life and industry. 

Unlike image classification and sementic segmentation tasks, multi-object detection has a variable number of bounding boxes as targets. In this aspect, the fundamental problem in training a multi-object detection network is, \textit{``How should the network learn a variable number of bounding boxes in different input images?''}

As an answer to this question, instead of directly modeling a variable number of bounding boxes, methods have been developed to learn a variable number of bounding boxes by discretizing the bounding box space and directly assigning the ground truth to the network's output. These methods have become the mainstream of training multi-object detection networks.

\begin{figure*}[t]
\begin{center}
\includegraphics[width=0.98\linewidth]{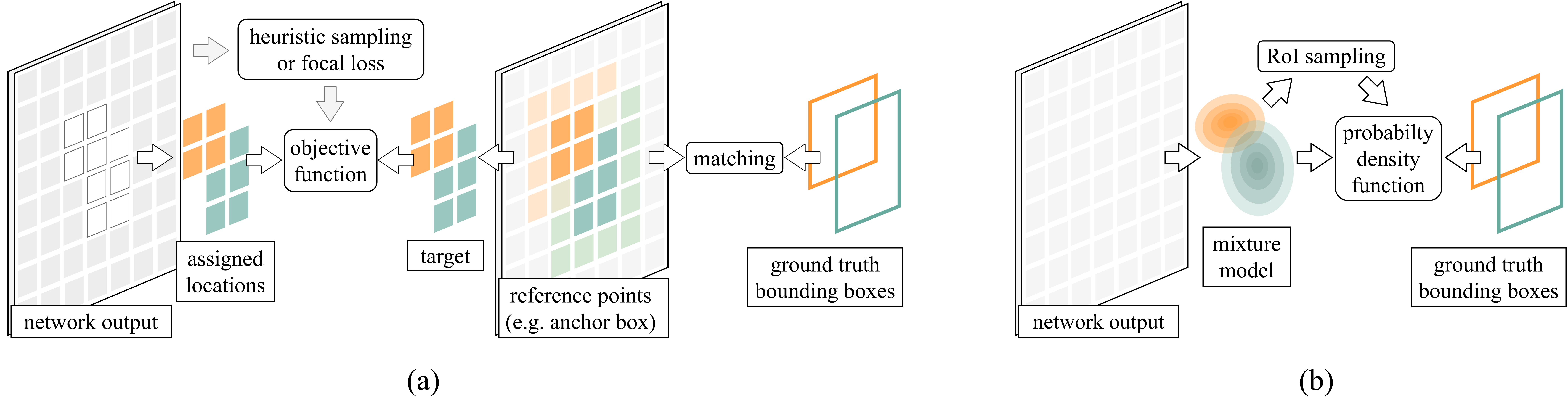}
\end{center}
\vspace{-3.5mm}
\caption{(a) The conventional training procedure that directly assigns the ground truths to specific locations of the network’s output.(b) Our proposed density estimation-based training that learns bounding box distribution.}
\vspace{-3.0mm}
\label{fig:detection-training}
\end{figure*}

Figure \ref{fig:detection-training} shows the training procedure of these methods. First, the matching algorithm compares each ground truth bounding box with each reference point such as an anchor box or a center, and determines whether they match or not. Second, as a result of the matching algorithm, the network output corresponding to the reference point location matched with the ground truth bounding box is extracted as the assignment location. Third, targets (e.g. displacements, classes) for each assigned location of output are generated. Finally, the assigned location is trained by the generated target through the objective function. Here, the unassigned locations are considered background areas, and are not trained by the coordinate of the ground truth bounding box.

However, in order to successfully train a multi-object detection network through this procedure, thoughtful consideration for each step is required.
First, in both the localization and the classification, the steps to assign each ground truth to the network's output are needed to learn the coordinates of the ground truth bounding box. In these steps, a matching algorithm and the reference points such as anchor boxes are important, since the ground truth are only assigned to the locations determined \jy{by} the matching result. 
Second, in the classification, there exists a severe imbalance between foreground and background. By the matching result, assigned locations are trained as corresponding ground truth classes regarded as foreground, but unassigned locations are trained as the background class. Generally, unassigned locations outnumber the assigned ones. This foreground-background imbalance problem makes training difficult. To alleviate this problem, separate process such as heuristic sampling \cite{liu2016ssd, shrivastava2016training} or focal loss \cite{lin2017focal} is required.
Third, some hyper-parameters and hand-crafted components make the detection performance sensitive. For example, design of anchor boxes \cite{ren2015fasterRCNN, redmon2017yolo9000}, matching algorithm \cite{zhang2020bridging}, and hyper-parameters of the focal loss \cite{lin2017focal} sensitively affect the detection performance. This sensitivity increases the cost of training a multi-object detection network.

In this paper, \textit{our goal is to propose a novel method to reduce the complex processing and heuristics for training multi-object detection network.} To this end, we reformulate the multi-object detection task as a density estimation of bounding boxes (See Fig. \ref{fig:detection-training}). Our proposed multi-object detection network, Mixture-Density-based Object Detector (MDOD), predicts the distribution of bounding boxes for an input image using a mixture model of components consisting of continuous (Cauchy) and discrete (categorical) probability distribution. For each component of the mixture model, the continuous Cauchy distribution is used to represent the distribution of the bounding box coordinates (left, top, right and bottom) and the categorical distribution is used to represent the class probability of that box. For localization, the MDOD is trained to maximize the log-likelihood of the estimated parameters for the mixture of Cauchy (MoC) given the ground truth bounding boxes of input images. For classification, to include the background class into the learning process, we propose to use region of interest (RoI) sampling for obtaining RoIs, but, the RoIs are stochastically sampled from the estimated MoC, and not obtained heuristically. The log-likelihood of the sampled RoIs is maximized instead of that of the ground truth.
The contributions of the proposed method are threefold as the following:
\vspace{2mm} \\
\noindent 1. Unlike the previous methods, we reformulate the multi-object detection task as a density estimation of bounding boxes for an input image. Through this novel approach, the complex processings and heuristics in the training of multi-object detection can be reduced.
\vspace{2mm} \\
\noindent 2. We estimate the density of bounding boxes using a mixture model consisting of continuous (for the location) and discrete (for the class) probability distribution. To this end, we propose a new network architecture, MDOD, and the objective function for it. 
\vspace{2mm} \\
\noindent 3. We measured the detection performance and speed of our proposed method on MS COCO. In some primary evaluation results, MDOD outperforms the previous detectors in both terms of detection performance and speed.

\section{Related Works}
In most modern multi-object detection methods, the ground truth bounding boxes must be assigned to the network's output based on reference points such as anchor boxes or center locations. Faster R-CNN \cite{ren2015fasterRCNN} attempts to represent the space in which a box can exist on an image as much as possible by using a large number of anchor boxes having various scales and aspect ratios. A ground truth bounding box is assigned to an anchor box if the intersection over union (IoU) between this anchor box and the ground truth bounding box is above a threshold. In later studies, the use of anchor boxes became a standard. \cite{liu2016ssd, fu2017dssd, redmon2016you}. To design an anchor box, most of methods inherit the shape heuristically found in previous studies. YOLOv2 \cite{redmon2017yolo9000} and YOLOv3 \cite{redmon2018yolov3} find the optimal anchor boxes through K-means clustering. However, a large number of anchors worsens the so-called foreground-background imbalance problem, since the unassigned background anchor boxes outnumber the assigned foreground ones, which makes training difficult. Also, a careful design of the anchor is required as the scale and aspect ratio of the anchor affect detection performance much. To alleviate the foreground-background imbalance problem, Hard negative mining \cite{liu2016ssd} and OHEM \cite{shrivastava2016training} sample the negative RoIs (Region of Interests) with a high loss. Focal Loss \cite{lin2017focal} tackles this problem by concentrating on the loss of hard examples. However, it has the hyperparameters that should be heuristically searched. 
Recently, studies not using anchors have been conducted. \cite{tian2019fcos} learn ground truth bounding boxes based on the center location instead of anchor boxes. \cite{law2018cornernet, duan2019centernet, zhou2019bottom} use the keypoint-based method used in pose estimation. They learn the keypoints of the bounding boxes in the form of heatmaps. However, these methods still directly assign the ground truths to specific locations of network's output and use focal loss to alleviate the foreground-background imbalance problem.

On other hand, there are studies dealing with matching algorithm. \cite{zhang2020bridging} argues that what is important is how to assign the ground truth bounding boxes, not the anchor box shapes, and proposes an adaptive method that automatically divides positive and negative samples. \cite{zhang2019freeanchor} points out that the IoU-based hand-crafted assignment is a problem. It learns the matching between a ground truth bounding box and an anchor through maximum likelihood estimation. However, this only learns matching weights and it still needs to construct the hand-crafted bag of anchors based on IoU.

In the previous studies, the concept of probability distribution in multi-object detection is mainly used to express the uncertainty of bounding box coordinates. For each predicted RoI ($roi^{k}$), \cite{he2019bounding} modeled a bounding box coordinate prediction ($b^{i}_{coord}$) as a Gaussian distribution to estimate $p(b^{i}_{coord}|roi^{k},image)$. \cite{choi2019gaussian} estimated the density of a specific bounding box coordinate for a specific anchor ($anchor^{k}$) as a Gaussian distribution: $p(b^{i}_{coord}|anchor^{k},image) \sim \mathcal{N}$.

In this paper, we perform multi-object detection by learning the distribution of bounding boxes ($b$) for an image using a mixture model, i.e. we estimate $p(b|image)$. Unlike the previous methods mentioned above, our MDOD does not require to directly assign the ground truth bounding boxes to the specific locations of the network's output.

\section{Problem Formulation: Mixture Model for Object Detection}
The bounding box $b$ can be represented as a vector consisting of four coordinates (position) $b_{p}$ for the location (left-top and right-bottom corners) and an one-hot vector $b_{c}$ for the object class. 
In the problem of multi-object detection, the conditional distribution of $b$ for an image may be multi-modal, depending on the number of objects in an image. Therefore, our object detection network must be able to capture the multi-modal distribution. We propose a new model MDOD that can estimate the multi-modal distribution by extending the mixture density network \cite{bishop1994mixture} for object detection. MDOD models the conditional distribution of $b$ for an image using a mixture model whose components consist of continous and discrete probability distribution, which \jy{respectively} represents the distribution of bounding box coordinates and the class probability. In this paper, we use the Cauchy distribution as a continuous distribution and the categorical distribution as a discrete distribution. 
The probability density function (pdf) of this mixture model is defined as follows:
\begin{equation} \label{eq:mixture_model}
\begin{aligned}
p(b|image) &=  \sum_{k=1}^{K} \pi_{k}
\mathcal{F}(b_{p}; \mu_{k}, \gamma_{k}) \mathcal{P}(b_{c}; p_k).
\end{aligned}
\end{equation}
Here, $\mathcal{F}$ denotes the pdf of Cauchy\footnote{$\mathcal{F}(x; \mu, \gamma) = \frac{1}{\pi} \frac{\gamma}{(x-\mu)^2 + \gamma^2}$, where $\mu$ is the location parameter and $\gamma$ is the scaling parameter.}, 
and $\mathcal{P}$ denotes the probability mass function (pmf) of categorical distribution. The parameters $\mu_{k}$, $\gamma_{k}$, and $\pi_{k}$ are the location, scale, and, mixing coefficient of the $k$-th component. The $C$-dimensional vector $p_k$ is the probability for $C$ classes. 
The Cauchy distribution represents the four-coordinates of the bounding box $b_{p} = \{b_{l}, b_{t}, b_{r}, b_{b}\}$. To prevent the model from being overly complicated, we assume that each dimension of the bounding box coordinates is independent from the others. Thus, the pdf of Cauchy for the bounding box coordinates can be factorized as follows:
\jy{
\begin{equation} \label{eq:gaussian}
\begin{aligned}
\mathcal{F}(b_{p}|image) &= 
\mathcal{F}(b_{l}; \mu_{k,l}, \gamma_{k,l}) \times 
\mathcal{F}(b_{t}; \mu_{k,t}, \gamma_{k,t}) \\ & \times 
\mathcal{F}(b_{r}; \mu_{k,r}, \gamma_{k,r}) \times 
\mathcal{F}(b_{b}; \mu_{k,b}, \gamma_{k,b}).
\end{aligned}
\end{equation}
}
The objective of the MDOD is to accurately estimate the parameters of the mixture model by maximizing the log-likelihood of the ground truth bounding box $b$, as follows:
\begin{equation} \label{eq:formulation}
\begin{aligned}
\theta &= \argmax_{\theta} \mathbb{E}_{b \sim p_{data}(b|image)} \log p(b|image; \theta).
\end{aligned}
\end{equation}
Here, $p_{data}(b|image)$ is the empirical distribution of $b$ for a given an input image and $\theta$ is the parameter vector that includes mixture parameters $(\mu_k, \gamma_k, \pi_k)$ and the class probability $p_k$.
\begin{figure}[t]
\begin{center}
\includegraphics[width=0.80\linewidth]{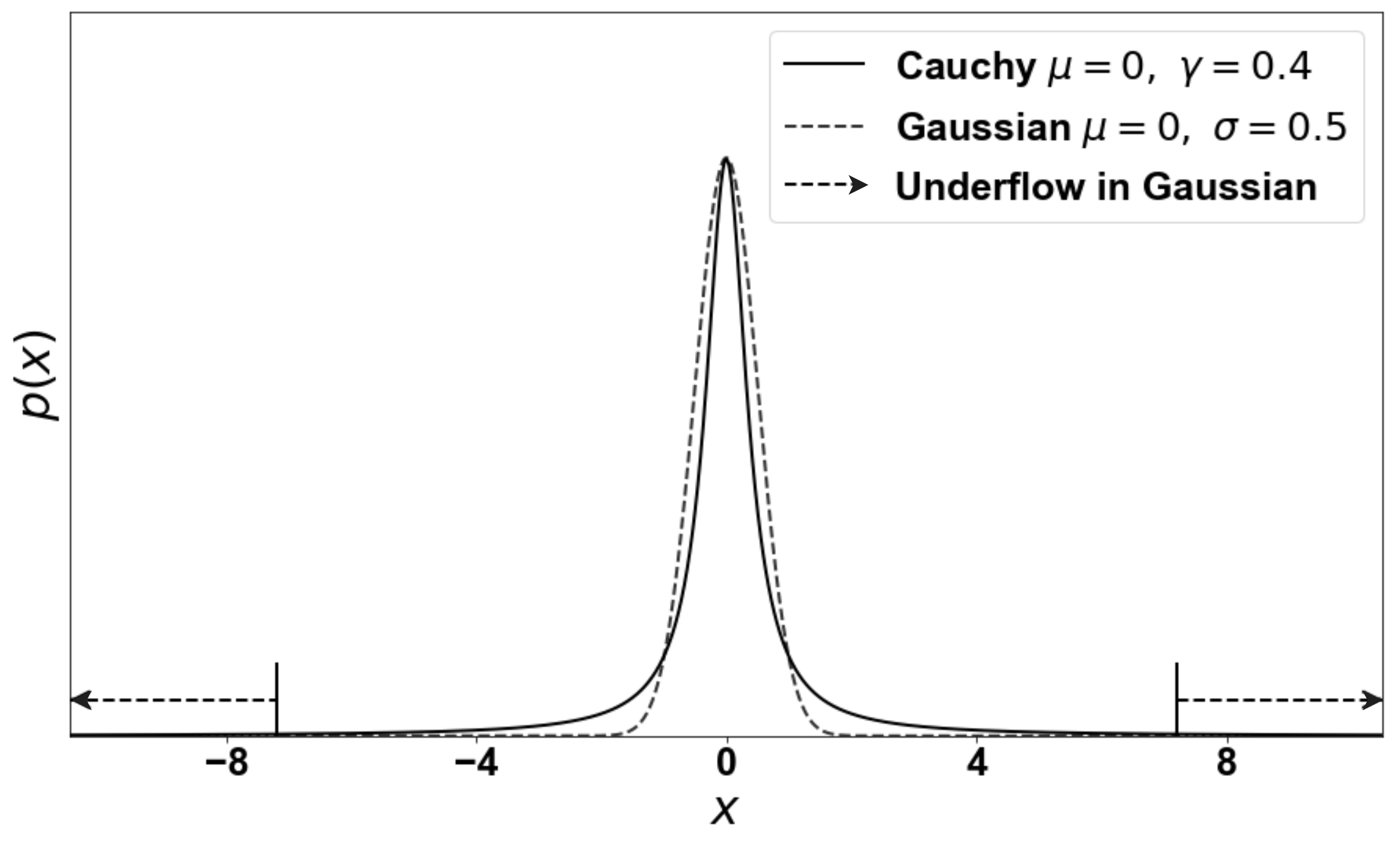}
\end{center}
\vspace{-5.0mm}
\caption{The pdfs of Gaussian and Cauchy distribution. Because of the limited precision of the floating point, for Gaussian, $p(x)=0$ for $|x| > 7.202$, i.e. underflow in log-likelihood calculation.}
\label{fig:underflow}
\end{figure}
\begin{figure*}[t]
\begin{center}
\includegraphics[width=0.90\linewidth]{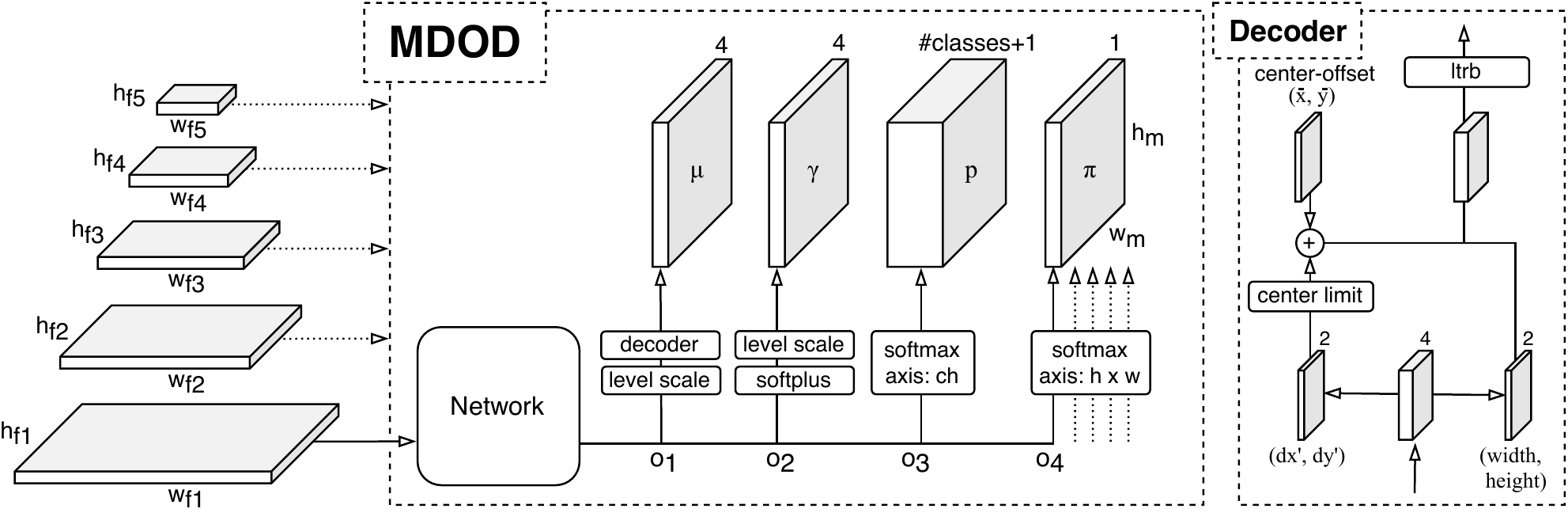}
\end{center}
\vspace{-3.5mm}
\caption{The architecture of MDOD. The parameters of the mixture model ($\mu$, $\gamma$, $p$, and $\pi$) are predicted by MDOD. The network produces its intermediate output $(o_1$ - $o_4)$ from each feature-map of the feature-pyramid.} 
\label{fig:architecture}
\vspace{-1mm}
\end{figure*}
\vspace{1mm} \\
\noindent \textbf{Cauchy vs. Gaussian: }
Gaussian distibution is one of the representative continous probability distribution. But, the likelihood of Gaussian distribution decreases exponentially as the distance from $\mu$ increases. Therefore, even if the predicted coordinate is slightly far away from the ground truth, underflow may arise due to the limited floating point precision in the actual implementation. It causes the problem that the likelihood becomes zero and the loss can not be backpropagated. On the other hand, as can be seen in Fig.~\ref{fig:underflow}, the Cauchy distribution has a heavier (quadratically decreasing) tail compared to the Gaussian distribution. Thus, there is \jy{much lesser chance} of the underflow problem.

\section{Mixture Density Object Detector (MDOD)}
\subsection{Architecture}
\label{sec:architecture}
Fig. \ref{fig:architecture} shows the architecture of MDOD. The network outputs $o_{1}$, $o_{2}$, $o_{3}$, and $o_{4}$ from the input feature-map. The parameter maps of our mixture model, $\mu$-map, $\gamma$-map, $p$-map, and $\pi$-map are obtained from $o_{1}$, $o_{2}$, $o_{3}$, and $o_{4}$, respectively. The mixture component is represented at each position on the spatial axis of the paramter-maps.

\begin{figure}[t]
\begin{center}
\includegraphics[width=0.73\linewidth]{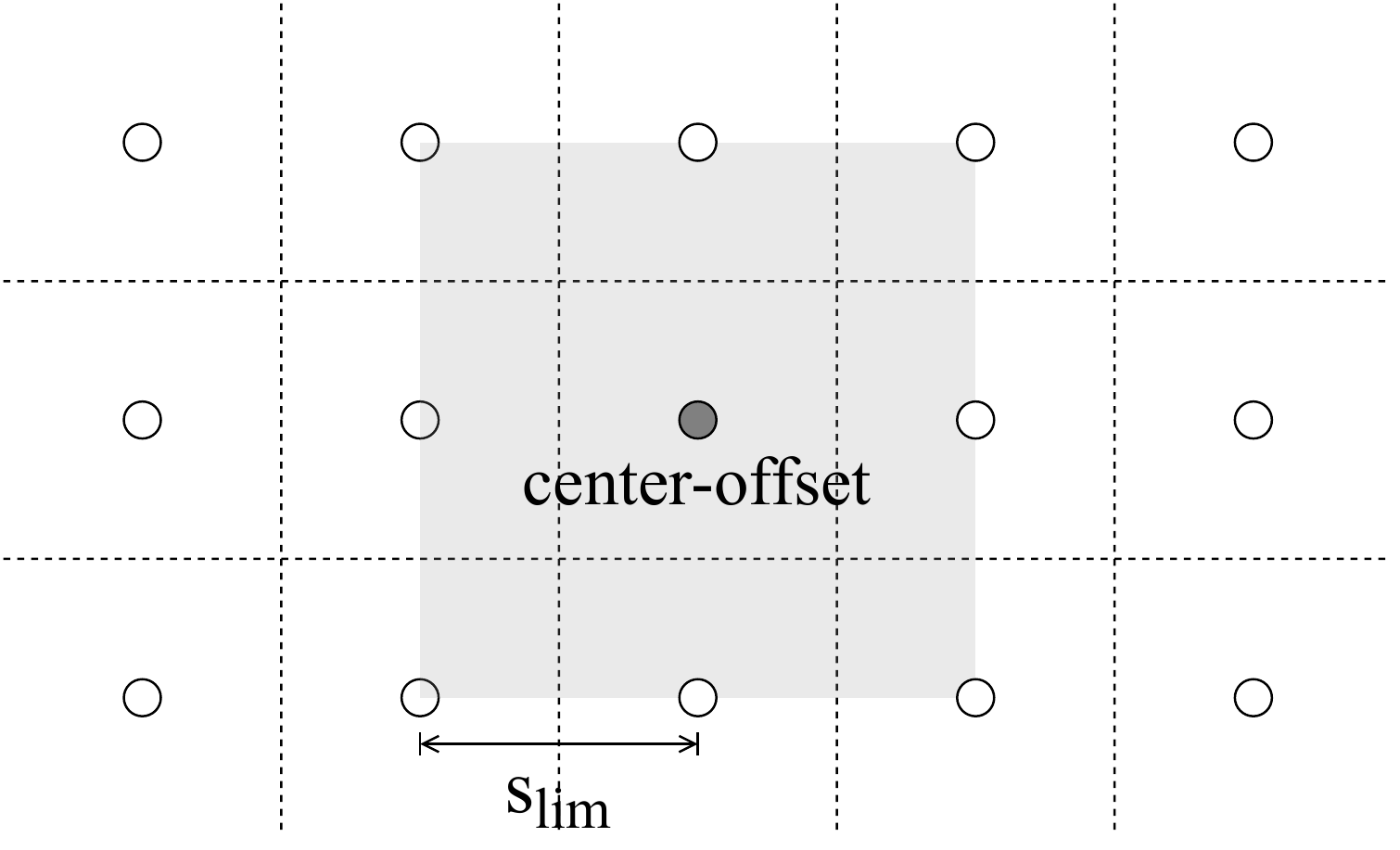}
\end{center}
\vspace{-4.5mm}
\caption{Illustration of the center-limit operation. The circles denote the center-offset. This operation limits $\mu_k$ within the gray area.}
\vspace{-3.5mm}
\label{fig:center_limit}
\end{figure}

The $\mu$-map is calculated from $o_{1} \in \mathbb{R}^{h_m \times w_m \times 4}$. First, each element of $o_1$ is scaled by a factor of $s = 2^{l-5}$ depending on the level $l \in \{1, \cdots, 5\}$ of the feature map in the feature pyramid as follows: $o_1' = s \times o_1$. Then, $o_1'$ is inputted to the decoder block. 
In the decoder block, the center-offset ($\bar{x}, \bar{y}$) is the default center position of the mixture components that are spatially aligned. Also, the center-limit operation illustrated in Fig. \ref{fig:center_limit} restricts the output not to deviate too much from the center-offset. \jy{This prevents the spatial misalignment of $\mu_{k}$.} It is implemented by applying $\tanh$ and multiplying the limit factor $s_{lim}$. In this paper, we set $s_{lim}$ equal to the spacing between adjacent center-offsets (see Fig. \ref{fig:center_limit}). 
The first two channels $(dx', dy')$ of $o_1'$ which correspond to the deviation from the center-offset are inputted to the center-limit operation. And then, the center-offset is added to the output of the center-limit operation. The overall computation of a center coordinate in $x$-direction is as follows: $x = \bar{x} + s_{lim} \times \tanh(dx')$. The same applies also to the $y$-direction. The last two channels of $o_1'$ act as the width and height. The $ltrb$-transformation converts coordinates represented by the center, width, and height ($xywh$) to the left-top and right-bottom corners ($ltrb$).

The $\gamma$-map is obtained by applying the softplus \cite{dugas2001incorporating} activation to $o_{2}$ and then multiplying the level-scale. The $p$-map is obtained by applying the softmax function along the channel axis to $o_{3} \in \mathbb{R}^{h_m \times w_m \times (C+1)}$, and the $\pi$-map is obtained by applying the softmax to the entire five spatial maps of $o_4\in \mathbb{R}^{h_m \times w_m \times 1}$ such that $\sum_{l=1}^5 \sum_{h=1}^{h_m^l} \sum_{w=1}^{w_m^l} \pi_{(h,w)}^l = 1$. Here, $C$ denotes the number of object classes and the last channel of $o_3$ is for the background class.

\jy{The network of MDOD consists of a 3$\times$3 kernel convolution layer and three 1$\times$1 kernel convolution layers.} Swish \cite{ramachandran2017searching} is used for the activation function of these layers except the output layer. We use 5-level Feature Pyramid Network (FPN) as a feature extractor \cite{lin2017feature}. MDOD estimates only one mixture model from all levels of feature-maps. Thus, the number of components $K$ is the summation of the number of components $(h_m \times w_m)$ of each parameter-map corresponding to the feature-map. Here, each feature-map $(o_1 - o_4)$ and the corresponding parameter-map ($\mu, \gamma, p, \pi$) at the same layer have the same dimension. 

\subsection{Training}
\label{sec:training}
\noindent \textbf{RoI sampling: }
To take the probability of the negative predictions into account, the class probabilities considering the background are commonly used as the confidence score of the predicted bounding box. But, while the input image has a background area, the set of ground truth bounding boxes $\{b_{gt}\}$ generally does not include the background class.
To obtain the bounding boxes of both foreground and background classes, we randomly sample the bounding box candidates from $\mu$ using $\pi$ which means the probabilities of the mixture components. 
If the IoU between a sampled candidate and a ground truth is above a threshold, we label it as the class of the ground truth with the highest IoU, otherwise, we label it as the background. Through this process, we create the RoI set $\{b_{roi}\}$. 

The $\{b_{roi}\}$ is sampled from $\mu$ and $\pi$ of the estimated MoC that are trained to represent the ground truth bounding box coordinates distribution for an input. Therefore, the foreground-background imbalance problem does not occur if the MoC estimates the distribution of bounding boxes well. The background bounding boxes in the $\{b_{roi}\}$ can be regarded as hard-negative samples. In the RoI sampling, these background samples are acquired stochastically, unlike the previous heuristic negative-mining methods \cite{liu2016ssd, shrivastava2016training}. In addition, since the RoI sampling is completely separated from the network structure, the structure of the network's output doesn't need to be considered. We need only apply the commonly used criterion (IoU$>$0.5) of the background.
\vspace{1mm} \\
\noindent \textbf{Loss function: }
For training MDOD to represent the background probability using $\{b_{roi}\}$, we define the loss function of MDOD into two terms. The first term is the negative log-likelihood of the MoC:
\begin{equation} \label{eq:mog_loss}
\begin{aligned}
\mathcal{L}_{MoC} = - \frac{1}{N_{gt}} \sum_{i=1}^{N_{gt}} \log \left( \sum_{k=1}^K \pi_k \mathcal{F}(b_{gt,p}^i ; \mu_k, \gamma_k) \right).
\end{aligned}
\end{equation}
Here, $(\pi_k, \mu_k, \gamma_k)$ depends on the image that contains the $i$-th ground truth bounding box $b_{gt}^i$. Note that Eq.(\ref{eq:mog_loss}) learns only the distribution of the coordinates of the ground truth bounding box $\{b_{gt,p}\} = \{b_{gt,p}^1, \cdots, b_{gt,p}^{N_{gt}}\}$, excluding class probability using the MoC parameters ($\pi, \mu, \gamma$). The second loss term is a complete form of our mixture model including class probability and is calculated as:
\begin{equation} \label{eq:cat_loss}
\begin{aligned}
\mathcal{L}_{MM} = - \frac{1}{N_{roi}} \sum_{j=1}^{N_{roi}} \log p(b_{roi}^j | image).
\end{aligned}
\end{equation}
$\mathcal{L}_{MM}$ is used to learn the class probability of the estimated mixture model. Note that $\mathcal{L}_{MM}$ is calculated using $\{b_{roi}\} = \{b_{roi}^1, \cdots, b_{roi}^{N_{roi}}\}$ sampled from $\mu$ and $\pi$ of the estimated MoC. Also, it is trained such that the MoC is not relearned by itself. To this end, the error is not propagated to other parameters of mixture models except class probabilities $p_k$. The final loss function is defined as:
\begin{equation} \label{eq:total_loss}
\begin{aligned}
\mathcal{L} = \mathcal{L}_{MoC} + \alpha \mathcal{L}_{MM}
\end{aligned}
\end{equation}
Here, $\alpha$ controls the balance between the two terms. In our experiments, we set $\alpha = 2$.

\subsection{Inference}
In the inference phase, we choose $\mu$’s of mixture components as coordinates of the predicted bounding boxes. We assume that these $\mu$’s have a high possibility to be close to the local maxima of the estimated mixture model. In the aspect of mixture-model-based clustering, we consider the $\mu$'s as representative values for the corresponding clusters. Before performing the non-maximum suppression (NMS), we can filter out the mixture components with relatively low $p_{c}$ or $\pi$ values. Since the scale of $\pi$ depends on the input image, we filter mixture components through normalized-$\pi$ ($\pi'$), which is obtained by normalizing $\pi$-vector with its maximum element, i.e. $\pi' = \pi/\max(\pi).$

\section{Experiments}
\subsection{Analysis for MDOD}
\begin{figure}[t]
\begin{center}
\includegraphics[width=0.78\linewidth]{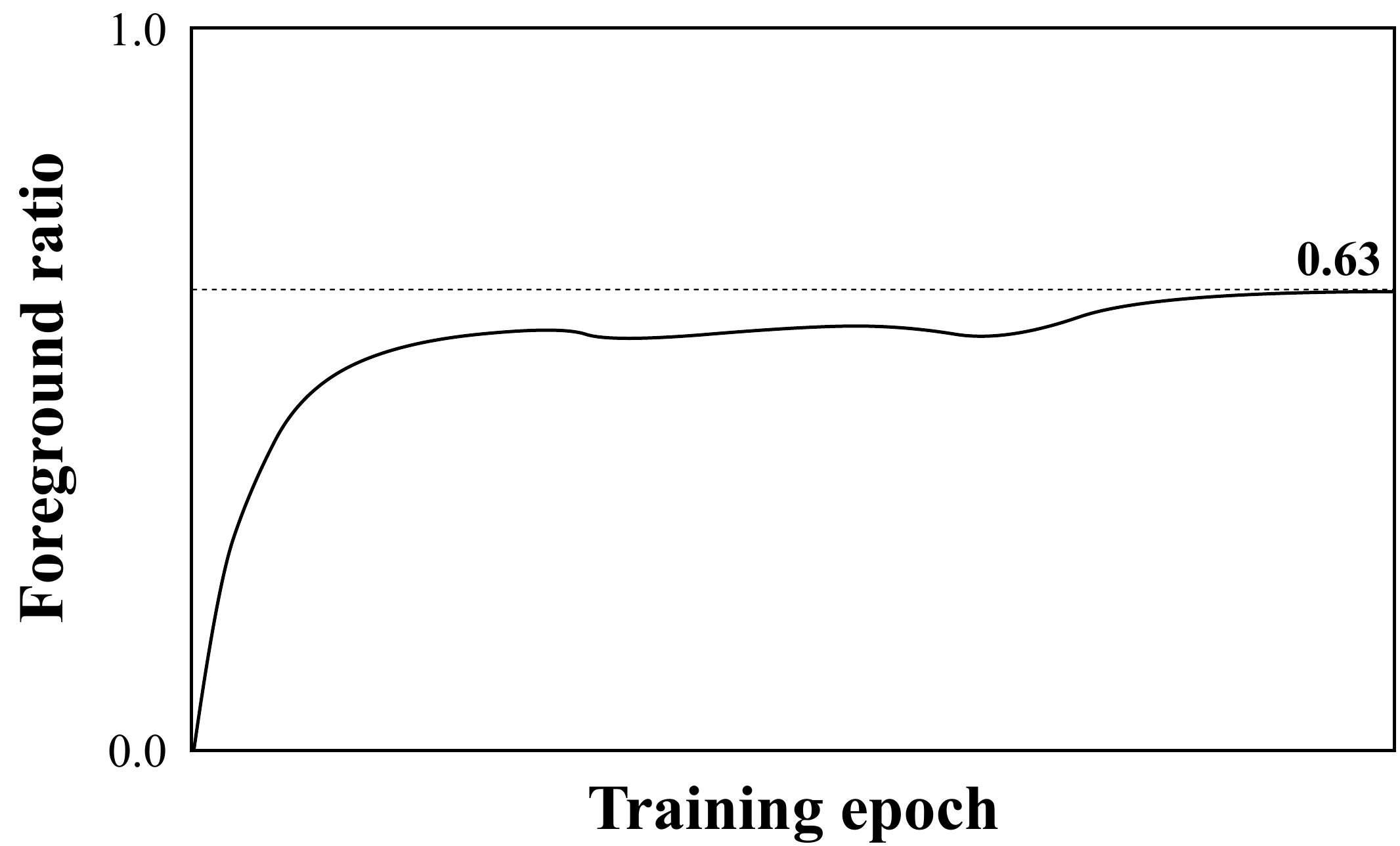}
\end{center}
\vspace{-3.0mm}
\caption{The ratio of foreground samples in the set $\{b_{roi}\}$ which is sampled from the mixture of Cauchy distribution at each training epoch.}
\label{fig:ratio_positive_negative}
\vspace{-3.0mm}
\end{figure}

\begin{figure}[t!]
\begin{center}
\includegraphics[width=0.78\linewidth]{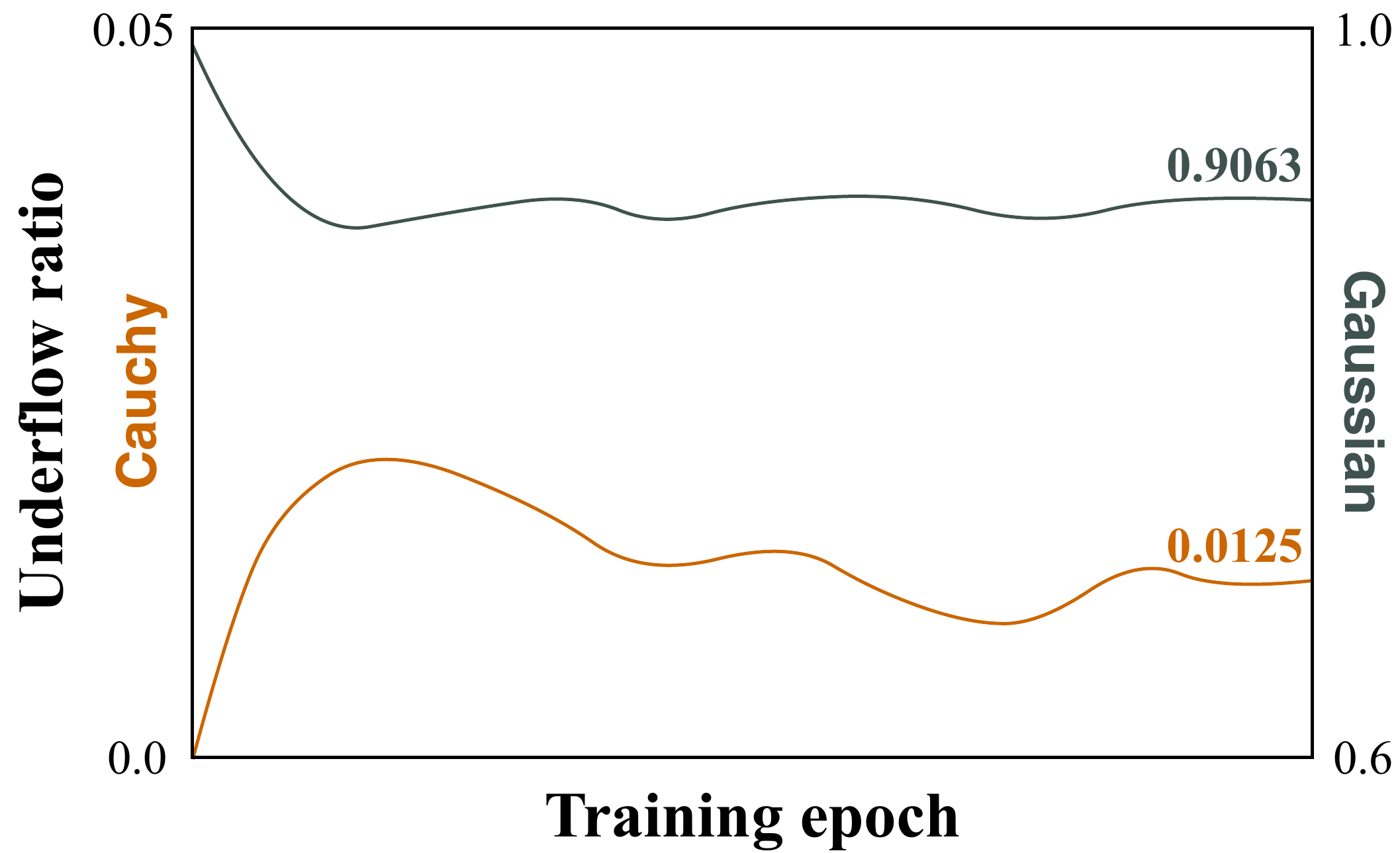}
\end{center}
\vspace{-3.0mm}
\caption{The ratios of underflowed components for Cauchy and Gaussian distributions at each training epoch.}
\label{fig:underflow_ratio}
\end{figure}

To analyze the MDOD, we use the MS COCO \cite{lin2014microsoft} `train2017' and `val2017' for training and evaluation. Input images are resized to 320$\times$320. ResNet50 \cite{he2016resnet} with FPN is used as feature extractor. Training details are described in the supplementary materials.
\vspace{1mm} \\
\noindent \textbf{Foreground-background balance: }
Since we perform sampling from the estimated MoC, the sampled $\{b_{roi}\}$ contains both foreground and background samples. In order to check the balance of foreground and background in $\{b_{roi}\}$, we measure the foreground ratio (\#foreground / \#total) of $\{b_{roi}\}$. In Fig. \ref{fig:ratio_positive_negative}, the foreground ratio is temporarily low at the initial of training but quickly increases as training progresses, and converges to a certain value. This shows that the foreground-background imbalance problem is solved naturally as the training progresses (\#foreground : \#background = 1.7 : 1 at the final epoch).
\vspace{1mm} \\
\noindent \textbf{Gaussian and Cauchy distribution: }
In practice, the likelihood of Gaussian and Cauchy distribution can be zero due to underflow caused by the limited floating-point precision. In order to show this problem during training, we measure the ratio of components where underflow occurs due to a large distance from a ground truth bounding box coordinate. As can be seen in Fig. \ref{fig:underflow_ratio}, in the Cauchy distribution, underflow rarely occurs, whereas in Gaussian, underflow occurs at a high ratio throughout the training process (about 0.9 ratio). The APs of MDOD using Gaussian and Cauchy distribution are 32.7 and 33.8, respectively.
\vspace{1mm} \\
\noindent \textbf{The number of RoIs: }
Table \ref{tab:n_sample} shows the APs changes according to $N_{roi}$, the size of $\{b_{roi}\}$. We set $N_{roi}$ proportional to $N_{gt}$, the number of the ground truth bounding boxes. As a result of the experiment, the performance is not sensitive to the $N_{roi}$. In this paper, $N_{roi}$ is set to three-times $N_{gt}$.
\vspace{1mm} \\
\noindent \textbf{Ablation study: }
MDOD has components that play a specific role in the intermidate feature-map. In this experiment, we change the following components in the MDOD architecture one by one to see the effect: \textit{ltrb-transformation (ltrb), center-limit} and \textit{level-scale} operation. Table~\ref{tab:ablation} shows the results. MDOD that uses all the components shows the best performance. Removing \textit{center-limit} and \textit{level-scale} operation results in a slight decrease in performance. The \textit{center-limit} and \textit{level-scale} operation seems to have a positive effect on detection results. If \textit{ltrb-transformation} is not used, bounding box is learned in $xywh$ coordinate. In our method, learning through the $ltrb$ coordinate shows around 1.0 better APs than learning through $xywh$.

\subsection{Evaluation result comparison}
\begin{figure*}[h!]
\begin{center}
\includegraphics[width=1.0\linewidth]{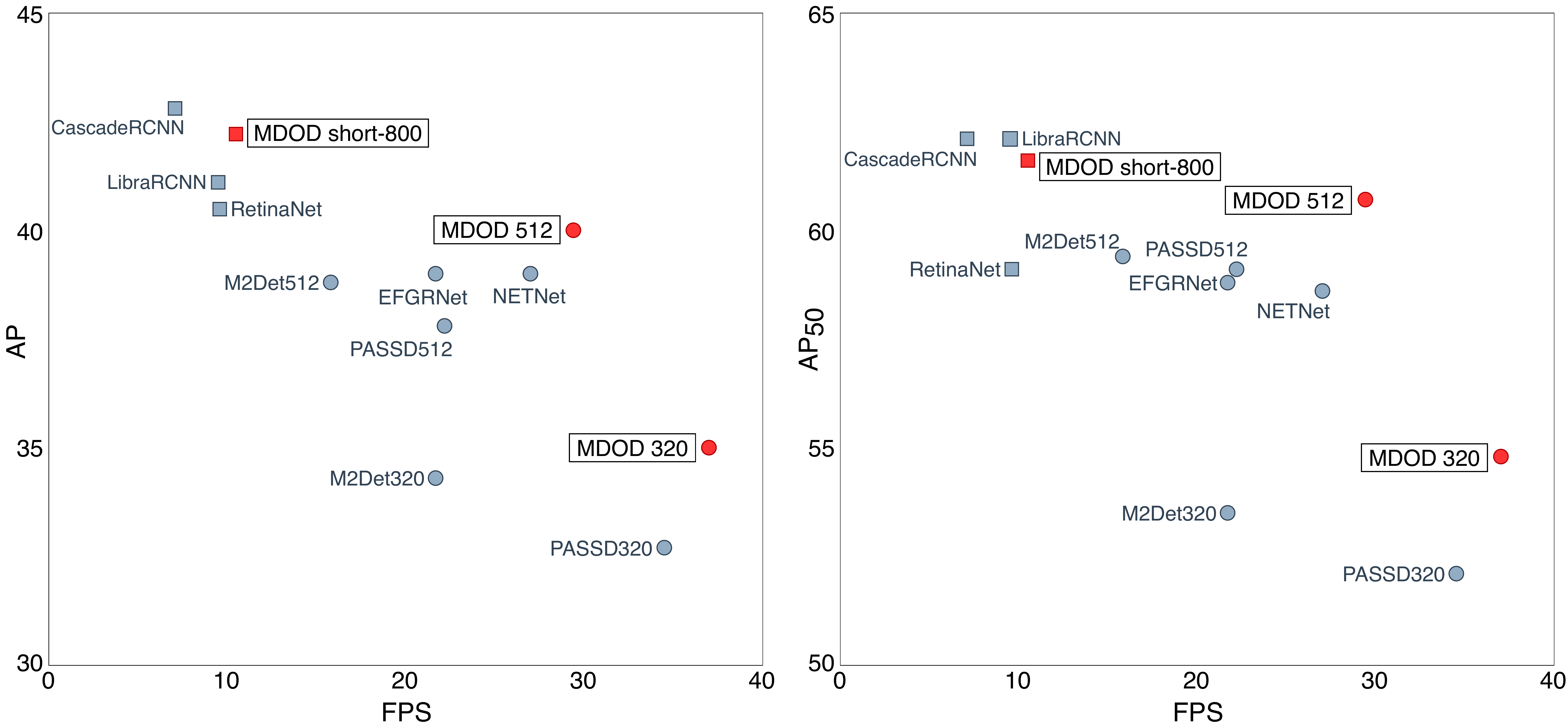}
\end{center}
\vspace{-3.0mm}
\caption{Comparison of speed (FPS) and APs. The circles and rectangles denote static and variable sized input, respectively.}
\label{fig:coco}
\end{figure*}

We compared MDOD with other object detection methods. MS COCO `train2017' dataset is used as the training-set and `test-dev2017' is used for evaluation. The frame-per-second (FPS) for MDOD is measured using a single nvidia Geforce 1080Ti including the post processing with batch size 1 without using tensorRT. Likewise, the FPSs for the other compared methods are also measured by the GPU with Nvidia Pascal architecture. Training details are described in the supplementary materials.
\vspace{1mm} \\
\noindent \textbf{Comparison with the baseline: }
We set up a simple baseline that learns bounding boxes through the conventional training method. In order to compare this baseline and MDOD as fairly as possible, we use the completely same batch size, augmentation strategy, and network architecture excluding the output layer to the baseline and MDOD. The baseline network is trained by smooth l1 and the cross entropy with hard negative mining. And, the baseline uses nine shapes of anchor boxes per each cell of output. 
\begin{table}[t!]
\small
\centering
\def\arraystretch{1.05}
\begin{tabular}{|M{2.0cm}|M{2.4cm}M{2.4cm}|}
\hline
$N_{roi}$         & $AP$ & $AP_{50}$ \\
\hline \hline
$N_{gt} \times 1$ & 33.8 & 53.3 \\
$N_{gt} \times 3$ & 33.8 & 53.4 \\
$N_{gt} \times 5$ & 33.9 & 53.3 \\
\hline
\end{tabular}
\caption{The size of $\{b_{roi}\}$ ($N_{roi}$) and APs.}
\label{tab:n_sample}
\end{table}
\begin{table}[t!]
\small
\centering
\def\arraystretch{1.05}
\begin{tabular}{|M{2.0cm}|M{1.0cm}M{1.0cm}M{1.0cm}M{1.0cm}|}
\hline
& \multicolumn{4}{c|}{MDOD} \\ 
\hline \hline
ltrb         & \checkmark & \checkmark & \checkmark & \\
center-limit & \checkmark & \checkmark &            & \checkmark \\
level-scale  & \checkmark &            &            & \checkmark \\ 
\hline
$AP$            & 33.8 & 32.9 & 32.3 & 32.8 \\
$AP_{50}$       & 53.4 & 52.9 & 51.6 & 52.5 \\ 
\hline
\end{tabular}
\caption{The effectiveness of the components of MDOD.}
\label{tab:ablation}
\vspace{-3.0mm}
\end{table}
As can be seen in the Table \ref{tab:coco}, MDOD outperforms the baseline. Also, in Table \ref{tab:speed}, MDOD shows a faster inference speed than the baseline. The reasons are as follows: The predictions of MDOD is only 1 for each cell in the output. Thus, the number of filters of output layer becomes smaller than that of the baseline (MDOD: 90, Baseline: 765). In addition, MDOD predicts fewer boxes than the Baselines (MDOD: 2134, Baseline: 19206). The number of predictions can affect the speed of the post-processing using NMS. 
\begin{table*}[t!]
\def\arraystretch{1.13}
\small
\begin{center}
\scalebox{0.95}{
    \begin{tabular}{|M{2.35cm}|M{2.35cm}|M{1.5cm}|M{0.9cm}M{0.9cm}M{0.9cm}|M{0.9cm}M{0.9cm}M{0.9cm}|}
    \hline
    method & feature extractor & input size & $AP$ & $AP_{50}$ & $AP_{75}$ & $AP_{S}$ & $AP_{M}$ & $AP_{L}$\\
    \hline \hline
    Baseline      & ResNet50-FPN   & 320x320   & 30.1  & 45.9  & 32.4  & 6.4   & 34.7  & 50.8 \\
    \textbf{MDOD} & ResNet50-FPN   & 320x320   & \textbf{33.9} & \textbf{53.8} & 35.5  & 14.7  & 35.1  & 49.6 \\ 
    Baseline      & ResNet101-FPN  & 320x320   & 31.1  & 46.8  & 33.6  & 6.7   & 36.1  & 52.3 \\
    \textbf{MDOD} & ResNet101-FPN  & 320x320   & \textbf{35.0} & \textbf{54.8} & 36.8  & 14.4  & 36.5  & 51.8 \\
    \hline
    Baseline      & ResNet50-FPN   & 512x512   & 35.0  & 53.2  & 38.1  & 15.0  & 40.2  & 50.7 \\
    \textbf{MDOD} & ResNet50-FPN   & 512x512   & \textbf{37.9} & \textbf{59.1} & 40.2  & 19.8  & 40.7  & 50.5 \\
    Baseline      & ResNet101-FPN  & 512x512   & 36.6  & 54.5  & 39.8  & 15.6  & 42.0  & 53.2 \\
    \textbf{MDOD} & ResNet101-FPN  & 512x512   & \textbf{40.0} & \textbf{60.7} & 42.6  & 20.7  & 43.1  & 53.8 \\
    \hline
    EfficientDet \cite{tan2020efficientdet} & Efficient-D0   & 512x512   & 33.8  & 52.2  & 35.8  & 12.0  & 38.3  & 51.2 \\
    \textbf{MDOD} & Efficient-D0   & 512x512   & \textbf{35.2} & \textbf{56.5} & 36.8  & 16.9  & 37.3  & 48.7 \\
    EfficientDet \cite{tan2020efficientdet} & Efficient-D1   & 640x640   & 39.6  & 58.6  & 42.3  & 17.9  & 44.3  & 56.0 \\
    \textbf{MDOD} & Efficient-D1   & 640x640   & \textbf{40.5} & \textbf{62.0} & 42.8  & 21.5  & 42.8  & 55.3 \\
    \hline
    \end{tabular}
}
\end{center}
\vspace{-3.5mm}
\caption{Evaluation result comparison of Baseline and EfficientDet with MDOD.}
\label{tab:coco}
\end{table*}
\begin{table*}[t!]
\def\arraystretch{1.13}
\small
\begin{center}
\scalebox{0.95}{
    \begin{tabular}{|M{2.3cm}|M{2.2cm}|M{1.5cm}|M{1.55cm}M{1.55cm}|M{1.55cm}|M{1.55cm}|}
    \hline
    method & feature extractor & input size & net-time & pp-time & total-time & FPS \\
    \hline \hline
    Baseline      & ResNet50-FPN   & 320x320   & 17 & 4 & 21 & 47.6 \\ 
    \textbf{MDOD} & ResNet50-FPN   & 320x320   & 16 & 2 & 18 & 55.6 \\ 
    Baseline      & ResNet50-FPN   & 512x512   & 22 & 6 & 28 & 37.5 \\ 
    \textbf{MDOD} & ResNet50-FPN   & 512x512   & 21 & 2 & 23 & 43.5 \\ 
    \hline
    \end{tabular}
}
\end{center}
\vspace{-3.5mm}
\caption{Inference time (ms) comparison of Baseline and MDOD. `net-time', `pp-time' and `total-time' mean network inference, post processing and total inference time, respectively.}
\label{tab:speed}
\vspace{0.0mm}
\end{table*}
\vspace{1.0mm} \\
\noindent \textbf{Comparison with EfficientDet: }
We compared the detection performance of MDOD with that of EfficientDet \cite{tan2020efficientdet}, a state-of-the-art method using the conventional training method. For the sake of fairness, the feature extractor used in EfficientDet is also applied to MDOD. In Table \ref{tab:coco}, this version of MDOD taking the structural superiority of EfficientDet's feature extractor shows better APs than the original EfficientDet in all the cases using the same feature extractor and input size. Especially, MDOD with Efficient-D1 achieved the highest AP (40.5) in this table. What is remarkable about these results is that this improvement is not caused by structural changes, heuristic or complex processing, but by a novel approach of learning distribution of bounding boxes in multi-object detection networks.
\vspace{1mm} \\
\noindent \textbf{Comparison with other methods: }
Tab. \ref{tab:coco_full} and Fig. \ref{fig:coco} show the APs and FPSs of object detectors. We compared MDOD with other representative methods using the similar feature extractor based on ResNet to focus on the methodology of multi-object detection. For comparison with more multi-object detection methods, we performed evaluations using not only static sized input images (320x320, 512x512) but also variable sized input (short-800). Here, the same augmentations used in RetinaNet, FCOS, and \textit{etc} are applied to train short-800 model.
In the comparison with the methods using a static sized input, MDOD clearly outperforms in both terms of detection performance and speed without any bells and whistles. The MDOD is not designed to speed up the inference time nor to reduce the computation. However, since MDOD has the advantages mentioned in the \textit{`comparison with the baseline'} and does not use modified convolution modules that require more computation, it shows faster inference speed than other methods when using the same input size and similar feature extractor. In comparison with the method using variable size input, MDOD shows promising results as a new detection methodology. But unlike the case of using a static sized input, it does not outperform the other state-of-the-art detectors. 
In the training with short-800 images, the number of mixture components $K$ is large and changed depending on the size of the input. Therefore, we speculate that large $K$, which changes during training, may interfere with the optimization. However, as the new approach to using mixture model-based density estimation, our method has a lot of room for advancement by further research for the mixture model and density estimation.

\section{Conclusion}
In this paper, we treat the multi-object detection task as a density estimation of bounding boxes for an input image. We proposed a new multi-object detector called as MDOD and the objective function to train it. MDOD captures the distribution of bounding boxes using the mixture model whose components consist of Cauchy and categorical distribution. Through this density-esimation-based approach and a new architecture of MDOD, we can reduce the complex precessing and heuristic for training multi-object detection network. In addition, we verified that the foreground-background imbalance problem is solved naturally as the training progresses in our method. We measured the detection performance and speed of MDOD on MS COCO. In the evalation using a static sized input, MDOD outperforms the other state-of-the-art multi-object detection methods in both terms of detection performance and speed. It is noteworthy that this performance is achieved not by structural changes or heuristic and complex processings, but by a new approach to multi-object detection. We believe that MDOD laid an initial step towards a new direction to multi-object detection which has a large room for improvements that can be achieved by further research and development.

\begin{table*}[t!]
\def\arraystretch{1.17}
\begin{center}
\scalebox{0.90}{
    \begin{tabular}{|M{3.5cm}|M{3.3cm}|M{2.2cm}|C{0.7cm}C{0.7cm}C{0.7cm}|C{0.7cm}C{0.7cm}C{0.7cm}|C{1.1cm}|}
    \hline
    method & feature extractor & input size & $AP$ & $AP_{50}$ & $AP_{75}$ & $AP_{S}$ & $AP_{M}$ & $AP_{L}$ & FPS   \\
    \hline \hline
    \multicolumn{10}{|l|}{\textit{static size input image:}} \\
    \hline
    SSD321 \cite{liu2016ssd, fu2017dssd}    & ResNet-101        & 321x321       & 28.0  & 45.4  & 29.3  & 6.2  & 28.3  & 49.3  & - \\
    RefineDet \cite{zhang2018single}        & ResNet-101 TCB    & 320x320       & 32.0  & 51.4  & 34.2  & 10.5 & 34.7  & 50.4  & - \\
    M2Det \cite{zhao2019m2det}              & ResNet-101 MLFPN  & 320x320       & 34.3  & 53.5  & 36.5  & 14.8  & 38.8  & 47.9  & 21.7 \\
    PASSD $\circ$ \cite{jang2019propose}    & ResNet-101 FPN    & 320x320       & 32.7  & 52.1  & 35.3  & 10.8  & 36.5  & 50.2  & 34.5 \\
    \textbf{MDOD} & ResNet-101 FPN          & 320x320           & \textbf{35.0}  & \textbf{54.8}  & 36.8  & 14.4  & 36.5  & 51.8  & \textbf{37.0} \\
    \hline
    SSD513 \cite{liu2016ssd, fu2017dssd} & ResNet-101 & 513x513         & 31.2  & 50.4  & 33.3  & 10.2 & 34.5  & 49.8  & - \\
    RefineDet \cite{zhang2018single} & ResNet-101 TCB & 512x512         & 36.4  & 57.5  & 39.5  & 16.6 & 39.9  & 51.4  & - \\
    M2Det \cite{zhao2019m2det} & ResNet-101 MLFPN & 512x512             & 38.8  & 59.4  & 41.7  & 20.5  & 43.9  & 53.4  & 15.8 \\
    PASSD $\circ$ \cite{jang2019propose} & ResNet-101 FPN & 512x512     & 37.8  & 59.1  & 41.4  & 19.3  & 42.6  & 51.0  & 22.2 \\
    EFGRNet \cite{Jing2019EFGR}     &  ResNet-101 & 512×512     & 39.0 & 58.8 & 42.3 & 17.8 & 43.6 & 54.5 & 21.7 \\
    NETNet \cite{li2020netnet}      & ResNet-101 NNFM & 512×512 & 38.5 & 58.6 & 41.3 & 19.0 &  42.3 & 53.9 & 27.0 \\
    \textbf{MDOD} & ResNet-101 FPN & 512x512      & \textbf{40.0}  & \textbf{60.7}  & 42.6  & 20.7  & 43.1  & 53.8  & \textbf{29.4} \\
    \hline \hline
    \multicolumn{10}{|l|}{\textit{variable size input image:}} \\
    \hline
    Faster R-CNN \cite{ren2015fasterRCNN} & ResNet-101 FPN & short-800  & 36.2  & 59.1  & 39.0  & 18.2  & 39.0  & 48.2  & -  \\
    Libra R-CNN \cite{pang2019libra} & ResNet-101 FPN & short-800       & 41.1  & \textbf{62.1} & 44.7  & 23.4  & 43.7  & 52.5  & 9.5 \\
    Cascade R-CNN \cite{cai2018cascade} & ResNet-101 FPN+ & short-800    & 42.8  & \textbf{62.1} & 46.3  & 23.7  & 45.5  & 55.2  & 7.1 \\
    RetinaNet \cite{lin2017focal}   & ResNet-101 FPN & short-800        & 39.1  & 59.1  & 42.3  & 21.8  & 42.7  & 50.2  & 9.6 \\
    FoveaBox \cite{kong2020foveabox}    & ResNet-101 FPN & short-800        & 40.8  & 61.4  & 44.0  & 24.1  & 45.3  & 53.2  & - \\
    FSAF \cite{zhu2019fsaf}     & ResNet-101 FPN & short-800        & 40.9  & 61.5  & 44.0  & 24.0  & 44.2  & 51.3  & - \\
    FCOS \cite{tian2019fcos}    & ResNet-101 FPN & short-800        & 41.5  & 60.7  & 45.0  & 24.4  & 44.8  & 51.6  & - \\
    ATSS \cite{zhang2020bridging}   & ResNet-101 FPN & short-800        & \textbf{43.6} & \textbf{62.1} & 47.4  & 26.1  & 47.0  & 53.6  & - \\  
    \textbf{MDOD} & ResNet-101 FPN  & short-800      & 42.2     & 61.6     & 45.1     & 25.3     & 44.6     & 51.7     & \textbf{10.5}    \\
    \hline
    \end{tabular}
}
\end{center}
\vspace{-3.0mm}
\caption{Evaluation results of various methods with MDOD. `$\circ$' denotes soft-nms \cite{bodla2017softnms}. The ‘short-800’ means to use an image that the shorter side is resized as 800 and the longer side is resized as smaller than 1333, while maintaining the aspect ratio.}
\label{tab:coco_full}
\vspace{3.0mm}
\end{table*}

{\small
\bibliographystyle{ieee_fullname}
\bibliography{egbib}
}

\end{document}